\pdfoutput=1
\documentclass[11pt]{article}
\usepackage[]{acl}
\usepackage{times}
\usepackage{latexsym}
\usepackage[T1]{fontenc}
\usepackage[utf8]{inputenc}
\usepackage{microtype}
\usepackage{graphicx}
\usepackage{tabularx}
\usepackage{booktabs}
\usepackage{nicefrac}
\usepackage{multirow}
\title{{D}ata-{E}fficient {A}utoregressive {D}ocument {R}etrieval for Fact Verification}

\author{James Thorne\thanks{\url{https://github.com/j6mes/sustainlp2022-deardr}}\\
        KAIST AI\\
        \texttt{thorne@kaist.ac.kr}}
\renewcommand\footnotemark{}
\begin{document}
\maketitle
\begin{abstract}
Document retrieval is a core component of many knowledge-intensive natural language processing task formulations such as fact verification and question answering. Sources of textual knowledge, such as Wikipedia articles, condition the generation of answers from the models. Recent advances in retrieval use sequence-to-sequence models to incrementally predict the title of the appropriate Wikipedia page given a query. However, this method requires supervision in the form of human annotation to label which Wikipedia pages contain appropriate context.
This paper introduces a distant-supervision method that does not require any annotation to train autoregressive retrievers that attain competitive R-Precision and Recall in a zero-shot setting.
Furthermore we show that with task-specific supervised fine-tuning, autoregressive retrieval performance for two Wikipedia-based fact verification tasks can approach or even exceed full supervision using less than $1/4$ of the annotated data indicating possible directions for data-efficient autoregressive retrieval.
\end{abstract}

\section{Introduction}
Conditioning answer generation on knowledge from textual sources is a common component of many well-studied natural language processing tasks. For example, in the SQuAD \citep{rajpurkar-etal-2016-squad} question answering task, a passage of text is used as a source of information to generate this answer. To enable machine-reading at scale, recent studies combine retrieval with reasoning \citep{chen-etal-2017-reading,roller-etal-2021-recipes} mandating that systems select appropriate passages from a corpus, such as Wikipedia, to condition answer generation.
Furthermore, tasks such as fact verification \citep{thorne-etal-2018-fever, wadden-etal-2020-fact,Diggelmann2020climate} use evidence retrieved from a corpus and consider both the label and the retrieved passages for evaluation.

Recent advances have been made in neural retrieval models, exploiting the structure of these tasks. \citet{de2020autoregressive} model retrieval as entity grounding \citep{bunescu-pasca-2006-using, le-titov-2018-improving}: the retriever is trained to predict the title of the Wikipedia document for a given input and is built on a seq2seq architecture. Even though GENRE \citep{de2020autoregressive} yields improvements for many retrieval-oriented NLP tasks in the KILT benchmark \citep{petroni-etal-2021-kilt}, the model requires supervision during training with labeled data that contains the document titles for a given input. 

\begin{figure}
    \centering
    \includegraphics[width=0.9\linewidth]{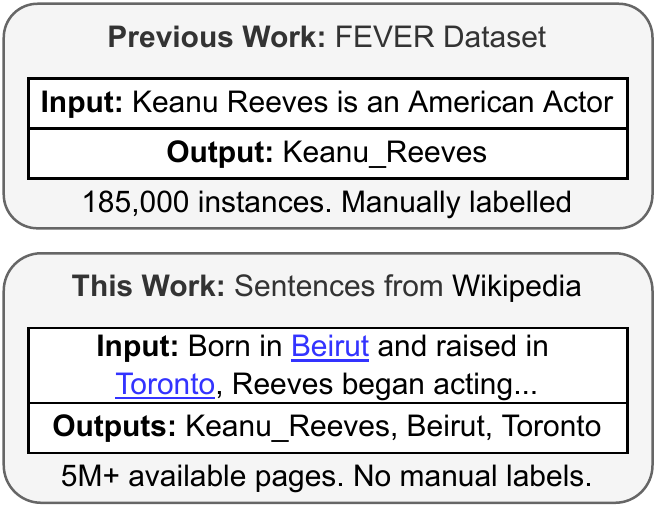}
    \caption{We present a distantly supervised pre-training objective for autoregressive information retrieval. Only using sampled sentences from Wikipedia, without labels, competitive scores can be attained for entity retrieval.}
    \label{fig:example}
\end{figure}


In this paper, we present a method for training a high-precision and high-recall retrieval system on Wikipedia data in a self-supervised manner. Our system can be trained in less than 6 hours on a single GPU without human-annotated data. The recall far exceeds conventional retrieval methods such as TF-IDF and BM25. Compared to GENRE \citep{de2020autoregressive}, which is trained with 11 annotated datasets for thousands of GPU-hours, our self-supervised approach for \textbf{D}ata \textbf{E}fficient \textbf{A}uto\textbf{R}egressive \textbf{D}ocument \textbf{R}etrieval, \textsc{DearDr}, performs within an R-Precision that is $6.36\%$ lower and a Recall@10 that is $0.91\%$ lower for the FEVER Shared Task. 
We additionally show that with task-specific fine-tuning, \textsc{DearDr} can attain precision and recall exceeding a fully supervised baseline for FEVER (and on-par with GENRE), with just 16K annotated instances rather than 109K in the full dataset. 
Similar findings are observed for the HoVer fact verification task \citep{jiang-etal-2020-hover}.

\section{Background}
Conventional information retrieval methods, such as TF-IDF and BM25, have been applied many knowledge-intensive NLP tasks \citep{chen-etal-2017-reading, thorne-etal-2018-fever}, with reasonable success. These methods do not require supervision: instead, query-document similarity is estimated based on token-level frequency information from observations on a fixed corpus. At test time, sparse discrete vector encodings of documents and the query are compared to return documents with the highest similarity to the query. While this aids application to new tasks and settings, recall can be low, especially when there are variations in phrasing due to the sparse encoding of which tokens (or variants considering n-grams or subtokens) are present.
Neural retrieval \citep{hanselowski-etal-2018-ukp, karpukhin-etal-2020-dense} in contrast, uses neural networks to generate dense encodings of the query and passages.
These models are trained with supervision: the training data contains lists of appropriate passages for a given query, but typically does not contain negative instances. 
How negative instances are sampled in training influences the suitability of the retrieved documents for the downstream task \citep{cohen2019learning, karpukhin-etal-2020-dense}. 
Variants of training regimes for dense retrieval also use Cloze-task \citep{lee-etal-2019-latent}, Wikipedia revision information \citep{chang2012link}, contrastive \citep{izacard2021towards} learning, or multi-task learning \citep{maillard-etal-2021-multi} to improve performance.

\subsection{Autoregressive Document Retrieval}
In contrast to the previous approaches, where the content of a passage is scored for a query using it's content, autoregressive document retrieval \citep{de2020autoregressive} uses a seq2seq model that is trained to predict a relevant document title, such as a Wikipedia page. Tokens are decoded incrementally left to right where the and scored with $p(\mathbf{y}|\mathbf{x}) = \prod_{i=1}p(y_i | y_{<i},\mathbf{x})$ where the decoded document title $\mathbf{\hat{y}} \in \mathcal{E}$ exists in a corpus $\mathcal{E}$.  To ensure this constraint is satisfied, constrained decoding sets $p(y_i | y_{<i,x},\mathbf{x}) = 0$ for token sequences $(y_1,\ldots\,y_i)$ that do not occur in the index. In practice this works well in the Wikipedia domain where document titles are simple canonical descriptors of an entity or concept.
An extension, mGENRE \citep{decao2020multilingual}, has been trained for multi-lingual entity linking using Wikipedia hyperlinks and internationalized versions of the pages from the Wikidata graph as supervision targets in other languages.
This has not been applied to an entity linking task, but not evaluated for document retrieval.

Similarly, GENRE did use hyperlink-based information by incorporating data from BLINK during training. However, its contribution to system performance appears low \citep[Table 8]{de2020autoregressive}, warranting further investigation. While the use of pre-training with hyperlink information in retrieval has shown promise \citep{ma2021pre} in other formulations, the use of distant-supervision in autoregressive retrieval, using the article titles and hyperlinks in training is emerging and has been studied in contemporaneous work \citep{chen2022corpusbrain}. \citet{lee2022generative} train autoregressive models for multi-hop retrieval, with a data augmentation strategy.  Alternative autoregressive retrieval formulations are designed to predict document sub-strings \citep{bevilacqua2022autoregressive}: this obviates the need to have unique document identifiers.



\section{Data Efficient Document Retrieval}
The primary objective of this paper is to reduce the dependency on supervised instances and exploit \textit{distant supervision} to train an autoregressive document retrieval system. Distant supervision for \textsc{DearDr} exploits the structural aspects of the Wikipedia corpus: specifically, the page titles (denoted PT) and hyperlinks (denoted HL) from sentences to other pages. Sentences from Wikipedia documents are sampled from the corpus as input and \textsc{DearDr} is trained to decode the page title or hyperlinks, or both (denoted PTHL).

Even though the \textsc{DearDr} has only been pre-trained with distant supervision without exposure to annotated training data for a knowledge intensive NLP task, we hypothesize that training to predict a Wikipedia page title or hyperlinks acts a a reasonable analog that simulates a common component of many retrieval oriented tasks.
This should be sufficient to allow zero-shot application to retrieve relevant documents without the need for human annotations enabling application of knowledge-intensive NLP tasks to new domains or languages. 
In contrast, the GENRE model \citep{de2020autoregressive} is trained with data from eleven Wikipedia-based NLP tasks with millions of annotated instances.



\paragraph{Constrained decoding} 
At test time, a result set is decoded by aggregating the results from a beam search \citep{sutskever2014bs} with constrained decoding \citep{de2020autoregressive}. 
However, in contrast to GENRE, which predicts a single entity per beam,  \textsc{DearDr} is trained to predict a sequence of all the hyperlinked page names (illustrated in Figure~\ref{fig:example}). 
 
\paragraph{Self-supervised vs task-specific retriever} Once the \textsc{DearDr} retriever has been pre-trained with self-supervision on Wikipedia data, it can be applied in a \textit{zero-shot} setting to the knowledge-intensive task of fact verification. 
Some aspects of the test-task formulation, may have patterns that differ to what \textsc{DearDr} is exposed to during the pre-training. 
Using small numbers of task-specific training data, the pre-trained \textsc{DearDr} will be fine-tuned evaluated on downstream tasks. We hypothesize that the pre-training regimen for \textsc{DearDr} will reduce the number of instances needed to train the system and attain similar performance to a system with full supervision.

\section{Experimental Setup}
Three different pre-training regimens for \textsc{DearDr}, based on page title (PT), sentence hyperlinks (HL) and a combination of both (PTHL), are performed using the snapshot of Wikipedia from June 2017. This was the snapshot used for the FEVER shared task. Document-level retrieval for two fact verification tasks will be evaluated: FEVER \citep{thorne-etal-2018-fever} and HoVER \citep{jiang-etal-2020-hover}.


\paragraph{Zero-Shot Document Retrieval:}
Without exposure to the underlying test task, \textsc{DearDR} will be pre-trained using unlabeled instances from English Wikipedia articles (pre-trained with PT, HL or PTHL), and then applied to instances from these retrieval-based NLP tasks. From Wikipedia, we generate 16.8M distant-supervision instances. 

\paragraph{Data-Efficient Document Retrieval:} The \textsc{DearDr} model will be fine-tuned using a low number of labeled instances from the target task. During training, we sample instances uniformly at random. We optimize training for \emph{Recall} with early stopping. This occurred after 12,500 steps (100K instances in total). 

\paragraph{Supervised Baseline:} For a controlled baseline system that \textsc{DearDr} can be compared against, we train a document retriever for the target task using all available data in a fully supervised setting. 

\paragraph{Previous Work:} We compare to GENRE \citep{de2020autoregressive} which was trained with data from \textit{eleven} retrieval tasks. We also compare to sparse-vector retrieval methods such as BM25 and TF-IDF. Finally, for dense-vector retrieval, we compare against DPR \citep{karpukhin-etal-2020-dense}. Because contrastive retrieval \citep{izacard2021towards} does not offer significant advantages over BM25 for FEVER, we do not evaluate against it.

\subsection{Evaluation}
Document retrieval is evaluated using two performance evaluation metrics: R-Precision and Recall@k. 
R-Precision is the precision of retrieved documents@R where R is the number of expected elements labeled for the instance. If the test set only specifies 1 valid document, this is equivalent to Precision@1. However, datasets sometimes require a multi-hop combination of pages for inference, requiring multiple documents to be considered for evaluation.
Recall@k is the proportion of the gold documents present in the first k elements predicted by the model. This metric is a useful indicator of potential upper-bound system performance for some tasks, such as FEVER, which considers up to 5 retrieval results for scoring claim veracity.  
Where multiple answer sets are present for instances, we consider each answer set independently and return the max score over all the sets to allow comparison to the KILT methodology \citep{petroni-etal-2021-kilt}.

\subsection{Implementation}
We use the HuggingFace \citep{wolf-etal-2020-transformers} implementation of T5-base \citep{2020t5}. This is fine-tuned using data as outlined in the previous section. We optimize hyper-parameters by sweeping the learning rate and scheduler (documented in Appendix~\ref{appendix:opt}) and maximizing R-Precision on the dev split. The index for constraining decoding is constructed from all subtokens  generated by the T5 Tokenizer for article titles from the Wikipedia version for the test task.


\section{Results and Discussion}

\subsection{Pre-training Intrinsic Evaluation}
DearDr was optimized by selecting the model with the highest R-precision on the FEVER shared task. For the R-Precision on the PT and HL components of the pre-training task, we provide the following intrinsic evaluation listed in Table \ref{tab:intrinsic} to evaluate the pre-training objectives.  Without pre-training on hyperlinks, recall is low indicating that hyperlink pre-training may be beneficial to multi-entity retrieval needed for some FEVER instances.
\begin{table}[t]
\centering
\begin{tabular}{@{}lcccc@{}}
\toprule
\multicolumn{1}{c}{\multirow{2}{*}{\textbf{Trainer}}} & \multicolumn{2}{c}{\textbf{R-Precision (\%)}} & \multicolumn{2}{c}{\textbf{Recall@10 (\%)}}\\ \cmidrule(l){2-5} 
\multicolumn{1}{c}{} & \multicolumn{1}{c}{\textbf{Page}} & \multicolumn{1}{c}{\textbf{Link}} & \textbf{Page} & \textbf{Link} \\ \midrule
\textit{GENRE} & \textit{26.51} & \textit{38.45} & \textit{36.28} & \textit{55.31}  \\ \midrule
PT & 33.12 & 21.10 & 40.04 & 29.95 \\
HL & 2.65 & \textbf{71.91} & 17.49 & 84.35 \\  
PTHL & \textbf{33.17} & 38.91 & \textbf{37.70} & \textbf{84.89} \\
\bottomrule
\end{tabular}
\caption{R-Precision and Recall of page titles (page), and hyperlink destinations (link) of sentences sampled from Wikipedia using our training approaches (PT, HL, PTHL) compared to a contemporary supervised approach which only underwent task-specific traiing and did not undergo the pre-training.}
\label{tab:intrinsic}
\end{table}

\subsection{Downstream Extrinsic Evaluation}
\paragraph{FEVER:} For the FEVER shared task, we trained \textsc{DearDr} with instances sampled from the Wikipedia snapshot for the task without using any human-annotated data. Table~\ref{tab:fever} highlights the retriever's R-Precision and Recall@10 in comparison to a fully supervised system showing that in the zero shot setting (without exposure to any labeled data) document retrieval scores are adequate and far exceed retrieval from token-based similarity methods such as TF-IDF and BM25.
Because FEVER is a claim verification task, the claims are similar in nature to sentences sampled from Wikipedia pages. The similarity between the claims and the Wikipedia sentences the zero-shot system was exposed to during training mean that this system is able to apply well to this task.

\begin{table}[t]
\begin{tabular}{lcc}
\toprule
\multicolumn{1}{c}{\multirow{2}{*}{\textbf{Approach}}} & \multicolumn{2}{c}{\textbf{FEVER Retrieval (\%)}} \\ \cmidrule(l){2-3} 
\multicolumn{1}{c}{} & \multicolumn{1}{c}{\textbf{R-Prec}} & \multicolumn{1}{c}{\textbf{Recall@10}} \\ \midrule
\textsc{DearDr} (PT) ZS & 77.66 & 91.95  \\
\textsc{DearDr} (HL) ZS & 56.55 & 89.94 \\
\textsc{DearDr} (PTHL) ZS &  75.89& 88.18  \\ \midrule
\textsc{DearDr} (PT) 16K  & 82.49 & \textbf{94.85} \\
Supervised & 81.36 & 94.28  \\ \midrule
GENRE (11 tasks) & \textbf{84.02} & 92.86 \\ 
TF-IDF &  29.89 & 68.57 \\
BM25  & 40.42 & 70.58 \\
DPR & 55.98 & 77.53 \\
\bottomrule
\end{tabular}
\caption{Without exposure to training instances from the FEVER task, \textsc{DearDr} attains high recall and R-Precision (R-Prec) for document retrieval for FEVER in a zero-shot (ZS) setting and can be further improved with fine-tuning on 16K (16,000) data.}
\label{tab:fever}
\end{table}

While GENRE was trained on 11 tasks with over 100K fact verification instances and over 500K question answering instances, the R-Precision of our zero-shot system is only 6.36\% lower with Recall@10 less than one percent lower. Given that most modern fact verification approaches perform supervised re-ranking of sentences from these documents, we do not foresee this lower precision having such a large impact on the final task score.

\begin{figure}[t]
    \centering
    \includegraphics[width=\linewidth]{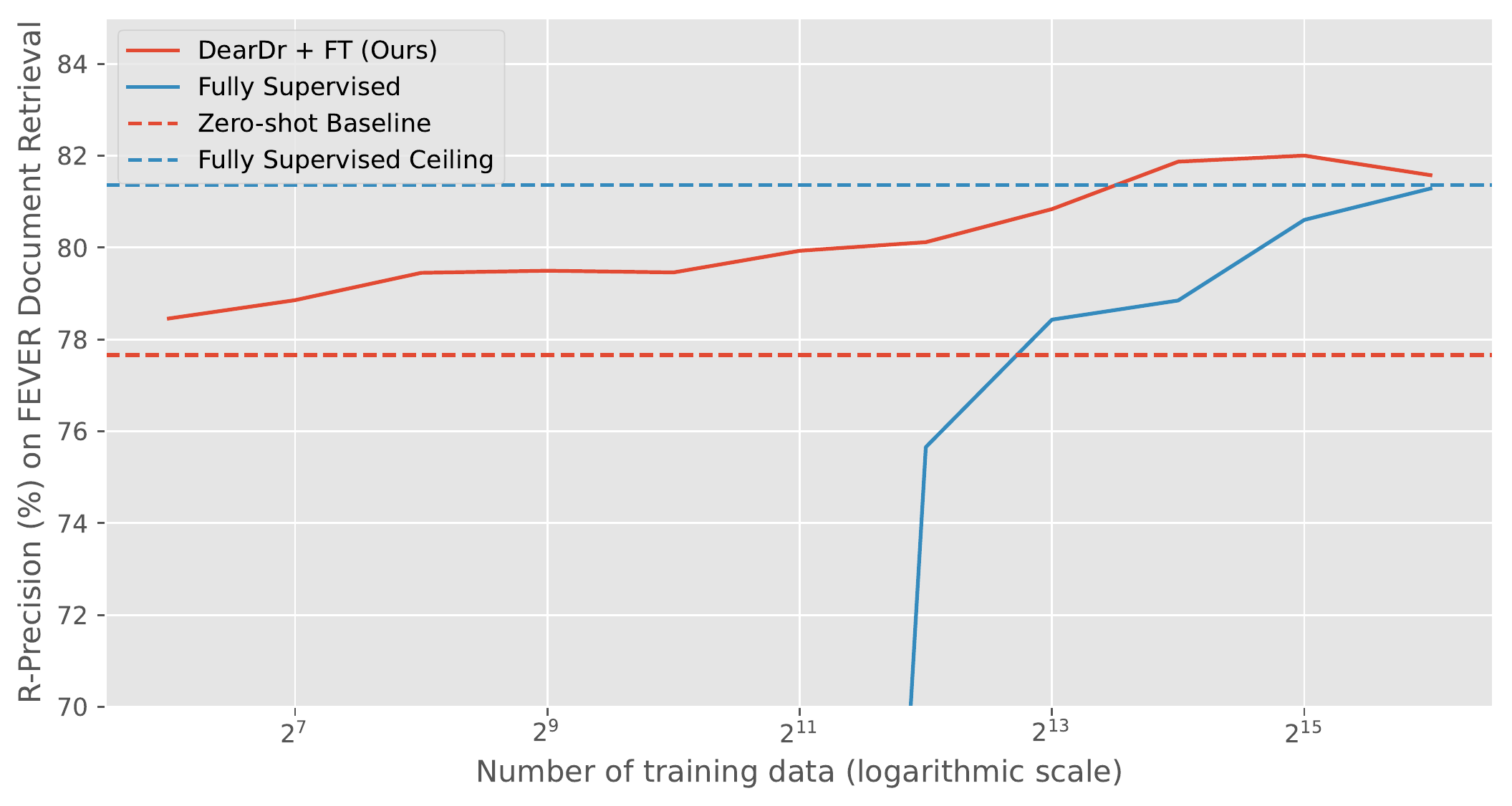}
    \caption{Learning curve showing greater R-Precision when training with fewer instances using \textsc{DearDr} pre-training compared to conventional supervised training}
    \label{fig:learning_curves}
\end{figure}

With full supervision from the FEVER task, our control model for comparison attains an R-Precision of 81.36\%.  
However, with small numbers of instances for fine-tuning on FEVER, higher recalls can be attained. With 16,384 instances (less than $1/8$ of the dataset), an R-Precision and Recall@10 exceeding this supervised baseline can be achieved. Furthermore, with only 2048 instances (2\% of the dataset), an R-Precision of at least 80\% is attained. 
In contrast, without pre-training R-Precision for both of these models with the same number of data is less than 30\%. Learning curves are plotted in Figure~\ref{fig:learning_curves}.

\paragraph{HOVER:} The multi-hop nature of HoVer presents more complex reasoning challenges than FEVER and is reported in Table~\ref{tab:hover}. Our zero-shot model has R-Precision that is less than 1\% of GENRE. While GENRE wasn't trained on HoVer, it was trained on HotpotQA \citep{yang-etal-2018-hotpotqa} which the HoVer dataset is derived from.

The benefit of pre-training with hyperlinks becomes apparent for multi-hop challenges as R-Precision for HL and PTHL exceed PT. With limited fine-tuning, using $1/4$ of the dataset, R-Precision with \textsc{DearDr} is less than 1\% away from a fully supervised model, despite using fewer data. Without pre-training, R-Precision is unsatisfactory. 
With all data, \textsc{DearDr} performs as good as a model without pre-training.
While \textsc{DearDr} is beneficial for this task with fewer data, there are clearly more complex challenges with multi-hop reasoning that require further data augmentation, such as \citep{lee2022generative},  to be solved solved by autoregressive methods for retrieval.

\begin{table}[t]
\centering
\begin{tabular}{lcc}
\toprule
\multicolumn{1}{c}{\multirow{2}{*}{\textbf{Approach}}} & \multicolumn{2}{c}{\textbf{HOVER Retrieval (\%)}} \\ \cmidrule(l){2-3} 
\multicolumn{1}{c}{} & \multicolumn{1}{c}{\textbf{R-Prec}} & \multicolumn{1}{c}{\textbf{Recall@10}} \\ \midrule
GENRE (Transfer) & {43.28} & {49.41} \\ \midrule
\textsc{DearDr} (PT) ZS & 32.83  & 36.43 \\
\textsc{DearDr} (HL) ZS & 42.22  & 43.78 \\
\textsc{DearDr} (PTHL) ZS& 38.94 & 47.44  \\ \midrule
\textsc{DearDr}(PTHL) 4K & 45.62 & 49.14  \\
\textsc{DearDr}(PTHL) All & \textbf{46.23} & 50.33  \\
Supervised 4K & 29.24 & 35.22 \\
Supervised & 46.22 & \textbf{50.38} \\
\bottomrule
\end{tabular}
\caption{The multi-hop aspect of HoVer presents new challenges. Despite this, DearDr attains higher R-Prec with fewer training data than supervised baselines}
\label{tab:hover}
\end{table}

\paragraph{Question Answering:} The similarity between fact verification and \textsc{DearDr} pre-training is similar, aiding retrieval. However, application to question answering (TriviaQA \citep[TQA]{joshi-etal-2017-triviaqa}, HotpotQA \citep[HPQA]{yang-etal-2018-hotpotqa} and NaturalQuestions \citep[NQ]{kwiatkowski-etal-2019-natural}) requires further study. Table~\ref{tab:qa} shows  pre-training does offer a benefit, but with fewer data for fine-tuning, similar gains in retrieval cannot be attained.  

\begin{table}[]
\begin{tabular}{@{}llll@{}}
\toprule
\multicolumn{1}{c}{\multirow{2}{*}{\textbf{Approach}}} &  \multicolumn{3}{c}{\textbf{R-Precision (\%)}}\\ \cmidrule(l){2-4} 
 & TQA  & HPQA & NQ \\ \midrule
GENRE* & 69.2 & 51.3 & 60.3 \\
\textsc{DearDr} (PTHL) ZS & 44.24 & 42.44 & 18.05 \\
\textsc{DearDr} (PTHL) 32K & 53.98 & 43.54 & 36.94 \\
\bottomrule
\end{tabular}
\caption{Application to question answering highlights further challenges (* reported by \citet{de2020autoregressive}).}
\label{tab:qa}
\end{table}

\section{Conclusions and Future Work}
We show that distant supervision and pre-training enables high precision autoregressive document retrieval with fewer annotated training data.  While previous work has studied the utility of pre-training for dense-retrieval, this work aids understanding of sparse autoregressive retrieval. In application to fact verification, fewer labeled training data were required. However, when we applied this method to question answering, satisfactory results were not obtained due to the domain shift between the two tasks. Better understanding this limitation would be required to adapt \textsc{DearDr} pre-training to a wider range of tasks and multi-hop reasoning.

\section*{Acknowledgments}
This work was supported by Institute of Information \& communications Technology Planning \& Evaluation (IITP) grant funded by the Korea government (MSIT)  (No.2019-0-00075, Artificial Intelligence Graduate School Program (KAIST)).

\label{sec:bibtex}

\bibliography{acl_anthology,custom}

\begin{thebibliography}{29}
\expandafter\ifx\csname natexlab\endcsname\relax\def\natexlab#1{#1}\fi

\bibitem[{Bevilacqua et~al.(2022)Bevilacqua, Ottaviano, Lewis, Yih, Riedel, and
  Petroni}]{bevilacqua2022autoregressive}
Michele Bevilacqua, Giuseppe Ottaviano, Patrick Lewis, Wen-tau Yih, Sebastian
  Riedel, and Fabio Petroni. 2022.
\newblock Autoregressive search engines: Generating substrings as document
  identifiers.
\newblock \emph{arXiv preprint arXiv:2204.10628}.

\bibitem[{Bunescu and Pa{\c{s}}ca(2006)}]{bunescu-pasca-2006-using}
Razvan Bunescu and Marius Pa{\c{s}}ca. 2006.
\newblock \href {https://aclanthology.org/E06-1002} {Using encyclopedic
  knowledge for named entity disambiguation}.
\newblock In \emph{11th Conference of the {E}uropean Chapter of the Association
  for Computational Linguistics}, pages 9--16, Trento, Italy. Association for
  Computational Linguistics.

\bibitem[{Cao et~al.(2021)Cao, Wu, Popat, Artetxe, Goyal, Plekhanov,
  Zettlemoyer, Cancedda, Riedel, and Petroni}]{decao2020multilingual}
Nicola~De Cao, Ledell Wu, Kashyap Popat, Mikel Artetxe, Naman Goyal, Mikhail
  Plekhanov, Luke Zettlemoyer, Nicola Cancedda, Sebastian Riedel, and Fabio
  Petroni. 2021.
\newblock \href {https://arxiv.org/abs/2103.12528} {Multilingual autoregressive
  entity linking}.
\newblock In \emph{arXiv pre-print 2103.12528}.

\bibitem[{Chang and Kao(2012)}]{chang2012link}
Yang-Jui Chang and Hung-Yu Kao. 2012.
\newblock Link prediction in a bipartite network using wikipedia revision
  information.
\newblock In \emph{2012 Conference on Technologies and Applications of
  Artificial Intelligence}, pages 50--55. IEEE.

\bibitem[{Chen et~al.(2017)Chen, Fisch, Weston, and
  Bordes}]{chen-etal-2017-reading}
Danqi Chen, Adam Fisch, Jason Weston, and Antoine Bordes. 2017.
\newblock \href {https://doi.org/10.18653/v1/P17-1171} {Reading {W}ikipedia to
  answer open-domain questions}.
\newblock In \emph{Proceedings of the 55th Annual Meeting of the Association
  for Computational Linguistics (Volume 1: Long Papers)}, pages 1870--1879,
  Vancouver, Canada. Association for Computational Linguistics.

\bibitem[{Chen et~al.(2022)Chen, Zhang, Guo, Liu, Fan, and
  Cheng}]{chen2022corpusbrain}
Jiangui Chen, Ruqing Zhang, Jiafeng Guo, Yiqun Liu, Yixing Fan, and Xueqi
  Cheng. 2022.
\newblock Corpusbrain: Pre-train a generative retrieval model for
  knowledge-intensive language tasks.
\newblock In \emph{Proceedings of the 31st ACM International Conference on
  Information \& Knowledge Management}, pages 191--200.

\bibitem[{Cohen et~al.(2019)Cohen, Jordan, and Croft}]{cohen2019learning}
Daniel Cohen, Scott~M. Jordan, and W.~Bruce Croft. 2019.
\newblock \href {https://doi.org/10.1145/3341981.3344220} {Learning a better
  negative sampling policy with deep neural networks for search}.
\newblock In \emph{Proceedings of the 2019 ACM SIGIR International Conference
  on Theory of Information Retrieval}, ICTIR '19, page 19–26, New York, NY,
  USA. Association for Computing Machinery.

\bibitem[{De~Cao et~al.(2020)De~Cao, Izacard, Riedel, and
  Petroni}]{de2020autoregressive}
Nicola De~Cao, Gautier Izacard, Sebastian Riedel, and Fabio Petroni. 2020.
\newblock Autoregressive entity retrieval.
\newblock In \emph{International Conference on Learning Representations}.

\bibitem[{Diggelmann et~al.(2020)Diggelmann, Boyd-Graber, Bulian, Ciaramita,
  and Leippold}]{Diggelmann2020climate}
Thomas Diggelmann, Jordan Boyd-Graber, Jannis Bulian, Massimiliano Ciaramita,
  and Markus Leippold. 2020.
\newblock \href {http://arxiv.org/abs/2012.00614} {Climate-fever: A dataset for
  verification of real-world climate claims}.
\newblock pages 1--16.

\bibitem[{Hanselowski et~al.(2018)Hanselowski, Zhang, Li, Sorokin, Schiller,
  Schulz, and Gurevych}]{hanselowski-etal-2018-ukp}
Andreas Hanselowski, Hao Zhang, Zile Li, Daniil Sorokin, Benjamin Schiller,
  Claudia Schulz, and Iryna Gurevych. 2018.
\newblock \href {https://doi.org/10.18653/v1/W18-5516} {{UKP}-athene:
  Multi-sentence textual entailment for claim verification}.
\newblock In \emph{Proceedings of the First Workshop on Fact Extraction and
  {VER}ification ({FEVER})}, pages 103--108, Brussels, Belgium. Association for
  Computational Linguistics.

\bibitem[{Izacard et~al.(2021)Izacard, Caron, Hosseini, Riedel, Bojanowski,
  Joulin, and Grave}]{izacard2021towards}
Gautier Izacard, Mathilde Caron, Lucas Hosseini, Sebastian Riedel, Piotr
  Bojanowski, Armand Joulin, and Edouard Grave. 2021.
\newblock Towards unsupervised dense information retrieval with contrastive
  learning.
\newblock \emph{arXiv preprint arXiv:2112.09118}.

\bibitem[{Jiang et~al.(2020)Jiang, Bordia, Zhong, Dognin, Singh, and
  Bansal}]{jiang-etal-2020-hover}
Yichen Jiang, Shikha Bordia, Zheng Zhong, Charles Dognin, Maneesh Singh, and
  Mohit Bansal. 2020.
\newblock \href {https://doi.org/10.18653/v1/2020.findings-emnlp.309}
  {{H}o{V}er: A dataset for many-hop fact extraction and claim verification}.
\newblock In \emph{Findings of the Association for Computational Linguistics:
  EMNLP 2020}, pages 3441--3460, Online. Association for Computational
  Linguistics.

\bibitem[{Joshi et~al.(2017)Joshi, Choi, Weld, and
  Zettlemoyer}]{joshi-etal-2017-triviaqa}
Mandar Joshi, Eunsol Choi, Daniel Weld, and Luke Zettlemoyer. 2017.
\newblock \href {https://doi.org/10.18653/v1/P17-1147} {{T}rivia{QA}: A large
  scale distantly supervised challenge dataset for reading comprehension}.
\newblock In \emph{Proceedings of the 55th Annual Meeting of the Association
  for Computational Linguistics (Volume 1: Long Papers)}, pages 1601--1611,
  Vancouver, Canada. Association for Computational Linguistics.

\bibitem[{Karpukhin et~al.(2020)Karpukhin, Oguz, Min, Lewis, Wu, Edunov, Chen,
  and Yih}]{karpukhin-etal-2020-dense}
Vladimir Karpukhin, Barlas Oguz, Sewon Min, Patrick Lewis, Ledell Wu, Sergey
  Edunov, Danqi Chen, and Wen-tau Yih. 2020.
\newblock \href {https://doi.org/10.18653/v1/2020.emnlp-main.550} {Dense
  passage retrieval for open-domain question answering}.
\newblock In \emph{Proceedings of the 2020 Conference on Empirical Methods in
  Natural Language Processing (EMNLP)}, pages 6769--6781, Online. Association
  for Computational Linguistics.

\bibitem[{Kwiatkowski et~al.(2019)Kwiatkowski, Palomaki, Redfield, Collins,
  Parikh, Alberti, Epstein, Polosukhin, Devlin, Lee, Toutanova, Jones, Kelcey,
  Chang, Dai, Uszkoreit, Le, and Petrov}]{kwiatkowski-etal-2019-natural}
Tom Kwiatkowski, Jennimaria Palomaki, Olivia Redfield, Michael Collins, Ankur
  Parikh, Chris Alberti, Danielle Epstein, Illia Polosukhin, Jacob Devlin,
  Kenton Lee, Kristina Toutanova, Llion Jones, Matthew Kelcey, Ming-Wei Chang,
  Andrew~M. Dai, Jakob Uszkoreit, Quoc Le, and Slav Petrov. 2019.
\newblock \href {https://doi.org/10.1162/tacl_a_00276} {Natural questions: A
  benchmark for question answering research}.
\newblock \emph{Transactions of the Association for Computational Linguistics},
  7:452--466.

\bibitem[{Le and Titov(2018)}]{le-titov-2018-improving}
Phong Le and Ivan Titov. 2018.
\newblock \href {https://doi.org/10.18653/v1/P18-1148} {Improving entity
  linking by modeling latent relations between mentions}.
\newblock In \emph{Proceedings of the 56th Annual Meeting of the Association
  for Computational Linguistics (Volume 1: Long Papers)}, pages 1595--1604,
  Melbourne, Australia. Association for Computational Linguistics.

\bibitem[{Lee et~al.(2022)Lee, Yang, Oh, and Seo}]{lee2022generative}
Hyunji Lee, Sohee Yang, Hanseok Oh, and Minjoon Seo. 2022.
\newblock Generative multi-hop retrieval.
\newblock \emph{arXiv preprint arXiv:2204.13596}.

\bibitem[{Lee et~al.(2019)Lee, Chang, and Toutanova}]{lee-etal-2019-latent}
Kenton Lee, Ming-Wei Chang, and Kristina Toutanova. 2019.
\newblock \href {https://doi.org/10.18653/v1/P19-1612} {Latent retrieval for
  weakly supervised open domain question answering}.
\newblock In \emph{Proceedings of the 57th Annual Meeting of the Association
  for Computational Linguistics}, pages 6086--6096, Florence, Italy.
  Association for Computational Linguistics.

\bibitem[{Ma et~al.(2021)Ma, Dou, Xu, Zhang, Jiang, Cao, and Wen}]{ma2021pre}
Zhengyi Ma, Zhicheng Dou, Wei Xu, Xinyu Zhang, Hao Jiang, Zhao Cao, and Ji-Rong
  Wen. 2021.
\newblock Pre-training for ad-hoc retrieval: hyperlink is also you need.
\newblock In \emph{Proceedings of the 30th ACM International Conference on
  Information \& Knowledge Management}, pages 1212--1221.

\bibitem[{Maillard et~al.(2021)Maillard, Karpukhin, Petroni, Yih, Oguz,
  Stoyanov, and Ghosh}]{maillard-etal-2021-multi}
Jean Maillard, Vladimir Karpukhin, Fabio Petroni, Wen-tau Yih, Barlas Oguz,
  Veselin Stoyanov, and Gargi Ghosh. 2021.
\newblock \href {https://doi.org/10.18653/v1/2021.acl-long.89} {Multi-task
  retrieval for knowledge-intensive tasks}.
\newblock In \emph{Proceedings of the 59th Annual Meeting of the Association
  for Computational Linguistics and the 11th International Joint Conference on
  Natural Language Processing (Volume 1: Long Papers)}, pages 1098--1111,
  Online. Association for Computational Linguistics.

\bibitem[{Petroni et~al.(2021)Petroni, Piktus, Fan, Lewis, Yazdani, De~Cao,
  Thorne, Jernite, Karpukhin, Maillard, Plachouras, Rockt{\"a}schel, and
  Riedel}]{petroni-etal-2021-kilt}
Fabio Petroni, Aleksandra Piktus, Angela Fan, Patrick Lewis, Majid Yazdani,
  Nicola De~Cao, James Thorne, Yacine Jernite, Vladimir Karpukhin, Jean
  Maillard, Vassilis Plachouras, Tim Rockt{\"a}schel, and Sebastian Riedel.
  2021.
\newblock \href {https://doi.org/10.18653/v1/2021.naacl-main.200} {{KILT}: a
  benchmark for knowledge intensive language tasks}.
\newblock In \emph{Proceedings of the 2021 Conference of the North American
  Chapter of the Association for Computational Linguistics: Human Language
  Technologies}, pages 2523--2544, Online. Association for Computational
  Linguistics.

\bibitem[{Raffel et~al.(2020)Raffel, Shazeer, Roberts, Lee, Narang, Matena,
  Zhou, Li, and Liu}]{2020t5}
Colin Raffel, Noam Shazeer, Adam Roberts, Katherine Lee, Sharan Narang, Michael
  Matena, Yanqi Zhou, Wei Li, and Peter~J. Liu. 2020.
\newblock \href {http://jmlr.org/papers/v21/20-074.html} {Exploring the limits
  of transfer learning with a unified text-to-text transformer}.
\newblock \emph{Journal of Machine Learning Research}, 21(140):1--67.

\bibitem[{Rajpurkar et~al.(2016)Rajpurkar, Zhang, Lopyrev, and
  Liang}]{rajpurkar-etal-2016-squad}
Pranav Rajpurkar, Jian Zhang, Konstantin Lopyrev, and Percy Liang. 2016.
\newblock \href {https://doi.org/10.18653/v1/D16-1264} {{SQ}u{AD}: 100,000+
  questions for machine comprehension of text}.
\newblock In \emph{Proceedings of the 2016 Conference on Empirical Methods in
  Natural Language Processing}, pages 2383--2392, Austin, Texas. Association
  for Computational Linguistics.

\bibitem[{Roller et~al.(2021)Roller, Dinan, Goyal, Ju, Williamson, Liu, Xu,
  Ott, Smith, Boureau, and Weston}]{roller-etal-2021-recipes}
Stephen Roller, Emily Dinan, Naman Goyal, Da~Ju, Mary Williamson, Yinhan Liu,
  Jing Xu, Myle Ott, Eric~Michael Smith, Y-Lan Boureau, and Jason Weston. 2021.
\newblock \href {https://doi.org/10.18653/v1/2021.eacl-main.24} {Recipes for
  building an open-domain chatbot}.
\newblock In \emph{Proceedings of the 16th Conference of the European Chapter
  of the Association for Computational Linguistics: Main Volume}, pages
  300--325, Online. Association for Computational Linguistics.

\bibitem[{Sutskever et~al.(2014)Sutskever, Vinyals, and Le}]{sutskever2014bs}
Ilya Sutskever, Oriol Vinyals, and Quoc~V Le. 2014.
\newblock \href
  {https://proceedings.neurips.cc/paper/2014/file/a14ac55a4f27472c5d894ec1c3c743d2-Paper.pdf}
  {Sequence to sequence learning with neural networks}.
\newblock In \emph{Advances in Neural Information Processing Systems},
  volume~27. Curran Associates, Inc.

\bibitem[{Thorne et~al.(2018)Thorne, Vlachos, Christodoulopoulos, and
  Mittal}]{thorne-etal-2018-fever}
James Thorne, Andreas Vlachos, Christos Christodoulopoulos, and Arpit Mittal.
  2018.
\newblock \href {https://doi.org/10.18653/v1/N18-1074} {{FEVER}: a large-scale
  dataset for fact extraction and {VER}ification}.
\newblock In \emph{Proceedings of the 2018 Conference of the North {A}merican
  Chapter of the Association for Computational Linguistics: Human Language
  Technologies, Volume 1 (Long Papers)}, pages 809--819, New Orleans,
  Louisiana. Association for Computational Linguistics.

\bibitem[{Wadden et~al.(2020)Wadden, Lin, Lo, Wang, van Zuylen, Cohan, and
  Hajishirzi}]{wadden-etal-2020-fact}
David Wadden, Shanchuan Lin, Kyle Lo, Lucy~Lu Wang, Madeleine van Zuylen, Arman
  Cohan, and Hannaneh Hajishirzi. 2020.
\newblock \href {https://doi.org/10.18653/v1/2020.emnlp-main.609} {Fact or
  fiction: Verifying scientific claims}.
\newblock In \emph{Proceedings of the 2020 Conference on Empirical Methods in
  Natural Language Processing (EMNLP)}, pages 7534--7550, Online. Association
  for Computational Linguistics.

\bibitem[{Wolf et~al.(2020)Wolf, Debut, Sanh, Chaumond, Delangue, Moi, Cistac,
  Rault, Louf, Funtowicz, Davison, Shleifer, von Platen, Ma, Jernite, Plu, Xu,
  Le~Scao, Gugger, Drame, Lhoest, and Rush}]{wolf-etal-2020-transformers}
Thomas Wolf, Lysandre Debut, Victor Sanh, Julien Chaumond, Clement Delangue,
  Anthony Moi, Pierric Cistac, Tim Rault, Remi Louf, Morgan Funtowicz, Joe
  Davison, Sam Shleifer, Patrick von Platen, Clara Ma, Yacine Jernite, Julien
  Plu, Canwen Xu, Teven Le~Scao, Sylvain Gugger, Mariama Drame, Quentin Lhoest,
  and Alexander Rush. 2020.
\newblock \href {https://doi.org/10.18653/v1/2020.emnlp-demos.6} {Transformers:
  State-of-the-art natural language processing}.
\newblock In \emph{Proceedings of the 2020 Conference on Empirical Methods in
  Natural Language Processing: System Demonstrations}, pages 38--45, Online.
  Association for Computational Linguistics.

\bibitem[{Yang et~al.(2018)Yang, Qi, Zhang, Bengio, Cohen, Salakhutdinov, and
  Manning}]{yang-etal-2018-hotpotqa}
Zhilin Yang, Peng Qi, Saizheng Zhang, Yoshua Bengio, William Cohen, Ruslan
  Salakhutdinov, and Christopher~D. Manning. 2018.
\newblock \href {https://doi.org/10.18653/v1/D18-1259} {{H}otpot{QA}: A dataset
  for diverse, explainable multi-hop question answering}.
\newblock In \emph{Proceedings of the 2018 Conference on Empirical Methods in
  Natural Language Processing}, pages 2369--2380, Brussels, Belgium.
  Association for Computational Linguistics.

\end{thebibliography}
\bibliographystyle{acl_natbib}

\appendix

\clearpage
\section{Appendix}

\subsection{Hardware Requirements}
Experiments were performed on a single workstation with a single NVIDIA GTX 1080 Ti GPU.

\subsection{Implementation}
\label{appendix:opt}
\subsubsection{Pre-training}
The following parameters were adjusted as part of the hyper-parameter optimization with the best parameters for all experiments indicated in bold. For HL, using learning rate of 5e-6 was more beneficial.

\begin{itemize}
\item Learning rate: 1e-4, \textbf{5e-5}, 1e-5, \textit{5e-6}, 1e-6.
\item Scheduler: \textbf{constant with warmup}, constant, linear.
\end{itemize}

\subsubsection{Fine-tuning + Supervised}
The following parameters were adjusted as part of the hyper-parameter optimization with the best parameters for all experiments indicated in bold. For fine-tuning, using a dropout of 0.2 was more beneficial. 

\begin{itemize}
\item Learning rate: 1e-4, 5e-5, \textbf{1e-5,} 5e-6, 1e-6.
\item Scheduler: \textbf{constant with warmup}, constant, linear.
\item Dropout: \textbf{0.1}, 0.2, 0.3

\end{itemize}




\section{Licenses}
The dataset released with this paper makes use of data from Wikipedia which is licensed under creative commons CC-BY-SA 4.0 license.

\end{document}